# An Expert System to Diagnose Spinal Disorders

Seyed Mohammad Sadegh Dashti [1]

*Department of Computer Engineering, Kerman Branch, Islamic Azad University, Kerman, Iran*

Seyedeh Fatemeh Dashti [2]

*Bushehr University of Medical Sciences, Bushehr, Iran*

**Abstract**

**Objective:** Until now, traditional invasive approaches have been the only means being leveraged to diagnose spinal disorders. Traditional manual diagnostics require a high workload, and diagnostic errors are likely to occur due to the prolonged work of physicians. In this research, we develop an expert system based on a hybrid inference algorithm and comprehensive integrated knowledge for assisting the experts in fast and high-quality diagnosis of spinal disorders.

**Methods:** First, for each spinal anomaly, the accurate and integrated knowledge was acquired from related experts and resources. Second, based on probability distributions and dependencies between symptoms of each anomaly, a unique numerical value known as certainty effect value was assigned to each symptom. Third, a new hybrid inference algorithm was designed to obtain excellent performance which was an incorporation of the Backward Chaining Inference and Theory of Uncertainty.

**Results:** The proposed expert system was evaluated in two different phases, real-world samples, and medical records evaluation. Evaluations show that in terms of real-world samples analysis the system achieved excellent accuracy. Application of the system on the sample with anomalies revealed the degree of severity of disorders and the risk of development of abnormalities in unhealthy and healthy patients. In the case of medical records analysis, our expert system proved to have promising performance, which was very close to those of experts.

**Conclusions:** Evaluations suggest that the proposed expert system provides promising performance, helping specialists to validate the accuracy and integrity of their diagnosis. It can also serve as an intelligent educational software for medical students to gain familiarity with spinal disorder diagnosis process, and related symptoms.

\* Corresponding author: Seyed Mohammad Sadegh Dashti
*E-mail address:* dashti@iauk.ac.ir

## 1. Background

The development and expansion of information technology have brought about new and essential functions in computer-based decision-making systems. Expert systems (ES), as an ingredient of artificial intelligence, have a central role in these systems. Decisions in an ES are made via computers. In these systems, experts' knowledge in a particular field of science is transferred to a computer, as a result of which experts' way of thinking in that particular area can be emulated. More specifically, the ES detects the logical models through which an expert makes a decision, making judgments similar to those of human beings [1, 2]. One of the specialized topics in physiotherapy and physical education is the detection of abnormalities. Spinal disorders are usually caused by some behaviors, bad habits, or some diseases. The spine, as one of the most vital parts of the human body, continually changes, and in some cases, it may experience anomalies due to its structure and the spinal arrangements [3]. One of the most severe anomalies involves a deformation of the spine and the upper body [4]. The side effects of these anomalies include scoliosis, flat back, round back (thoracic kyphosis), swayback (lumbar lordosis), and cervical lordosis [3]. Nevertheless, no medical ES has been implemented or designed to detect and prevent spinal disorders. This study suggests an ES designed to simulate the knowledge of experts and helps patients detect and prevent their harmful postural habits that could, in some cases, lead to premature death. A medical ES is composed of some features that distinguish it from other conventional medical systems. One of the significant differences in these systems as far as obtaining results is concerned, is that they can emulate doctors' reasoning and inference [5, 6]. A medical ES also helps to overcome some of the shortcomings that a human expert may display. The most critical shortcomings include outdated information, possible mistakes, geographical limitations, boredom, lack of quick or complete response time, unclear diagnostic processes, lack of flexibility, low accuracy, slow learning, and lack of varied expertise [5].

Patients at any age may develop spinal disorders. Spine and intervertebral discs are curved but have different structures. In a healthy person, typically, the spine has four curvatures, including cervical lordosis, thoracic kyphosis, lumbar lordosis, and sacral kyphosis [7]. These curves, in the bone and muscle structure of an average person, must be aligned at an acceptable angle. However, if they are deformed or experience abnormalities, they will cause anomalies and defects such as boredom, loss of movement functionality, respiratory restriction, reduced capacity of the heart, reduced blood circulation, and congenital disabilities.

Furthermore, due to the alternations imposed on the person's appearance, she/he may be psychologically vulnerable, while being exposed to the risk of increased injury and damage as a result of pressure engendered by abnormal curvature [8]. In the ES proposed in this study, five anomalies, namely scoliosis, flat back, round back, swayback (lumbar lordosis), and cervical lordosis, are examined. Two different theories form the basis of the proposed system, including backward chaining, the theory of uncertainty; each of them is described separately in the next chapters. The paper is organized as follows:

In section 2, the rationale behind this research is discussed. In section 3, elements of inference are explained. Studies in the literature are reviewed in section 4. In section 5, the methodology used in knowledge engineering is discussed. Section 6 describes the system evaluation investigated in this study. Finally, the paper presents the concluding remarks in section 7.

## 2. Rationale

Despite the development of medical science, industrial growth, and technology expansion in the modern world, there are still people threatened by new disorders, such as postural anomalies [9]. Spinal disorders which might be detected by the proposed expert system are described as follows:

- **Scoliosis:** a sideways curve in the spine is called "scoliosis" [10, 11, 12]. This anomaly is initially C-shaped but develops into an S-like shape in its advanced forms. Scoliosis can be detected by various methods [10, 11]. The disorder is occasionally very progressive. If the curve increases, due to the pressure it exerts on the lungs, it reduces lung capacity and respiratory functioning, as a result of which some patients experience a slow death [13, 14].
- **Flat back:** bounded curvature in the spine in the back or the waist, is called "flat back" [15]. In this anomaly, the upper body switches into a vertical position, and mobility in the spine decreases [15]. In the flat back, the cartilage and disc are damaged, and the defensive mechanism against imposed blows and spine resiliency both disappear [10].
- **Round back:** increased back curvature leads to a condition known as "round back" [16, 17]. Severe round back is of two types, namely the irreversible and the reversible [18, 19]. This anomaly disfigures the shape of the body, causing physiological effects and tightness in the chest. The tightness can disturb the activity of the cardio-pulmonary system [10].
- **Swayback:** an increase in the back curvature is called "swayback". In this anomaly, the curvature in the lumbar region increases [20]. The stomach area is pulled forward, usually accompanied by pain and fatigue in the back [21, 22]. Swayback leads to pressure on vertebral discs and supposedly engenders back pain, makes childbirth difficult in women, and causes erosion in the bones [10].
- **Cervical lordosis**: increased cervical curvature leads to a condition called "cervical lordosis." As a result of this anomaly, the individual experiences a forward head posture, in which and the height of the neck appears to be shorter than its standard height [10].

Failure to detect and treat these anomalies may leave harmful impacts on physiological functions, as in the case of the kyphosis effect, which disturbs the respiratory system [23]. In addition to physical damage, this adverse condition may even bring about psychological and social effects; for instance, kyphosis can be associated with depression [24].

Postural anomalies, apart from hereditary factors, are caused by industrialization, lack of exercise, bad eating habits, and the use of non-standard equipment [25]. In everyday life, an individual may experience different postural positions, some of which can lead to some anomalies in the long run [26]. Additionally, individuals who use pieces of equipment that do not conform to ergonomic standards, and those who ignore anthropometric dimensions are more likely to develop structural and physiological disorders [27]. Recent longitudinal studies show that the most common form of scoliosis is late-onset idiopathic scoliosis [12] that leads to a physical disorder, fatigue, and back pain. If left untreated, this disorder can even contribute to the mortality rate of the public population, because strenuous physical activity causes cardiac arrest over time [13, 14].

There are two significant motivations behind designing and implementing the proposed ES. First, helping experts to diagnose spinal anomalies more accurately; Second, determining the potential risk of development of spinal disorders in healthy people. Even more, since the integrity of knowledge is ensured in the development of the proposed system, it could be easily used by medical students to gain an understanding of the detection of spinal disorders and related symptoms.

## 3. Basic Elements

In this section. First, we discuss the technical architecture of expert systems. Next, fundamental theories and inference approaches used in the development of the proposed expert system are thoroughly explained to provide readers with the basics necessary to understand how the proposed approach works.

### 3.1 Rule-Based Expert System

In order to find solutions, conventional computer problem-solving programs use well-structured algorithms, data structures, and crisp reasoning strategies. In the face of the severe problems with which expert systems are concerned, it may be more useful to employ heuristics: strategies that often lead to a correct solution but sometimes fail. Conventional expert systems based on rules use human expertise to solve real-world problems that would typically require human intelligence. Knowledge of experts is often expressed on a computer in the form of rules or data. Based on the problem condition, specific rules and information can be retrieved to solve problems. Rule-based expert systems have played a significant role in strategic goal setting, planning, development, scheduling, fault control, diagnosis, and so on in modern intelligent systems and their implementations [28].

Today's users can choose from hundreds of commercial software packages with friendly graphical user interfaces with the technological advances made in the last decade [29].
. Conventional computer programs carry out tasks using a decision-making logic containing very little knowledge other than a basic algorithm to solve this particular problem. Basic knowledge is often incorporated as part of the programming code so that as knowledge changes, the program has to be updated. Small pieces of human knowledge are compiled from knowledge expert systems into a knowledge base; that is used to infer through a problem using appropriate knowledge. A significant benefit here is that using the same program without reprogramming attempts; a particular problem can be solved within the knowledge base domain. Besides, expert systems can explain the rationale process and tackle the level of confidence and ambiguity that traditional algorithms do not manage [30]. Some of the critical advantages of expert systems are as follows:
• Replication and maintenance of irreplaceable human experience;
• Ability to deliver a system which is more reliable than human experts in terms of consistency;
• Minimizing the need for human expert presence needed at several locations at the same time (especially in a dangerous environment that is hazardous to human health);
• solutions can be developed much faster when compared to the training procedure of human experts;

Figure 1 displays the essential components of the expert system. All relevant information, details, rules, and relations used by the experts are stored in the Knowledge Base. The knowledge base may incorporate many human experts ' expertise [28]. A rule is a conditional statement that links the premises to actions or results

Another method that is used to collect and store information in a knowledge base is the frame. It links an item or entity to a set of facts or values. frame-based representation of knowledge is a well-suited method for object-oriented programming techniques. Also referred to as are expert systems that use frames to store knowledge in the knowledge base are usually referred to as frame-based expert systems. The inference engine aims to search for knowledge base information and relationships and provide answers predictions and suggestions in the way a human expert might. The inference engine should find and compile the correct facts, definitions, and laws. Two types of inference approaches are widely used–backward chaining is the practice of beginning with hypotheses and moving backward to the facts that support them [31]. Forward chaining begins with the evidence and progresses to the conclusions [32]. The facility of explanation allows a user to understand how the expert system achieved those outcomes. The purpose of the knowledge acquisition facility is to provide an effective and secure means to collect and store all knowledge base components. Expert user interface software is very often used to model, upgrade, and use expert systems. The user interface's purpose is to make the use of an expert system simpler for designers, users, and administrators.

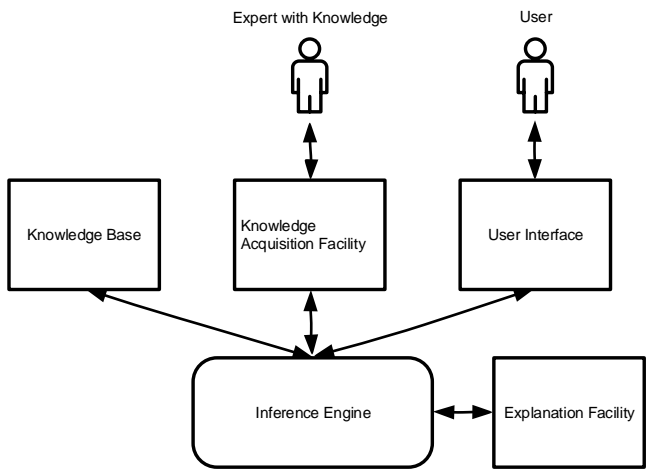

Fig. 1. Architecture of a typical expert system.

### 3.2 Inference in Rule-Based expert systems

A rule-based expert system is made up of if-then rules a set of facts, and an interpreter controlling the execution of the rules, fed with the facts. Such if-then rule statements are used to formulate the conditional statements which form the full base of knowledge. A single if-then rule presumes the form *'if x is A then y is B'* and if-part of rule *'x is A'* is referred to as the antecedent or premise while the following part of rule *' y is B'* is referred to as the consequence or conclusion for rule-based systems, there are two specific types of inference engines: forward chaining and backward chaining. The initial facts are first processed in a forward chaining system and continue to use the rules to draw new conclusions from those facts. The hypothesis (or solution/goal) we are trying to reach in a backward chaining system is processed first, and we continue to search for rules that would allow us to infer this hypothesis. New sub-goals are also set for testing as the processing progresses. Primarily, forward chaining systems are driven by data, whereas backward chaining systems are goal-driven. In circumstances where information is costly to obtain, forward chaining strategy is particularly appropriate. If there is not enough information about what the result of inference might be, or if there is some particular hypothesis to be validated, forward chaining systems may not be efficient. Backward chaining inference is useful in conditions where the amount of data is potentially very high and where it is of value to have some unique features of the system under consideration [33].

### 3.3 Backward Chaining Inference:

In the case of backward chaining, the primary concern is to align the conclusion of a rule against some known goal. So the' then' (consequent) part of the rule is generally not expressed as an action to be taken but rather as a condition, which is valid if the antecedent part(s) is correct. The backward chaining inference is analogous to the validation testing of the hypothesis in human problem-solving. For instance, a health-care specialist might suspect a patient's problems, which he/she then tries to prove by checking for specific symptoms. This reasoning style is designed by a goal-driven quest in an Expert System and is called backward-chaining [34], [35]. It is a theoretical top-down model that starts with a goal or hypothesis and searches for rules to validate the hypothesis. It tries to balance the variables that lead to relevant data facts and shows that the inference moves backward from the intended goal to establish facts that would fulfill the goal [36]. The implementation of backward chaining in a rule-based expert system is as follows:

1) the system checks the memory to see if the target has been added. Another knowledge base may have already proved the goal, so this phase is required. The algorithm reviews its set of rules, and if the goal has not been proved before, it continues to look for one or more that contains the goal in its THEN portion. This type of rule is called the goal rule.

2) Then the method looks at whether the target rule premises are listed in the memory. If the premises are not specified, new goals or sub-goals are to be checked, and other rules can be used to support them.

3) The system continues in this recursive way until it discovers a premise not provided by any rule; it is called a ' primitive'. The algorithm asks the user for details about it when a primitive is identified. This knowledge is then used by the system to prove both the sub-goals and the original goal.

**3.4 Certainty Factor:** Earlier rule-based expert systems assumed that all current information was either absolutely true or false. However, In the real world, there is often uncertainty associated with the set of rules and the data provided by the user [37].

Expert systems use the certainty factor (CF) to implement the theory of uncertainty, which produces a possible output value of certainty. Thus, knowledge engineers consider some degree of uncertainty generated by uncertain occurrences. This consideration is specifically essential in the diagnosis of some diseases with uncertain consequences [38]. The certainty value in this work is within the range of 0 to 100. In the proposed inference algorithm, certainty factor (CF)s are assigned both to the premises and antecedents. As the rule's premise is uncertain because of uncertain facts, and the conclusion is uncertain because of the rule's specification, the following formula is used to estimate the conclusion's certainty factor:

$$CF = (AntecedentCF) * (PremiseCF)/100 \qquad (1)$$

In some situations, there is more than one rule which supports a given conclusion. In this case, each of the rules can be fired and add to the CF. If a fact supports a conclusion and a rule fires for the favor of that conclusion, then the following

equations will be used to evaluate the fact-related current CF:

$$CF(X.Y) = X + \frac{Y(100-X)}{100} \quad (2)$$

$$\text{where } X \text{ and } Y > 0$$

$$CF(X.Y) = X + Y/(1 - min(|X|.|Y|))$$

$$\text{where } X \text{ or } Y < 0$$

$$CF(X.Y) = -CF(-X.-Y)$$

$$\text{where } X \text{ and } Y < 0$$

## 4. Literature Review

Until now, many different approaches have been introduced to analyze spine condition and diagnose anomalies and vertebrae defects. Detection of spinal anomalies is generally possible by non-invasive or invasive methods. Some invasive methods include X-ray images, fluoroscopic, CT, and MRI scans [39]. Non-invasive approaches fall into two categories. The first batch category includes the use of kyphometer, inclinometer, flexible ruler, spinal pantograph, electro-goniometer, spinal mouse. The second category, however, relies on non-contact methods, such as New York test and observation methods [40]. One of the oldest methods, which is considered to be a common practice among professionals, is Cobb's method. In addition to the risks of exposure to X-rays, time-consumption, and financial costs, this method is prone to errors caused by patients' physical movement during the imaging procedure. This invasive process can also cause bone cancer in men, as well as breast cancer and abortion in women in some cases [41]. The approach proposed by [16] is a good example of non-invasive methods. In 2010, [16] introduced a method for detecting anomalies of the spine, drawing on the image process, and the markers installed on the spinous process. They tested this method on forty male students of Birjand University. In this method, markers were installed on the naked body, and images of the individual's movement were recorded through the motion analyzer camera. Comparing their method with a flexible ruler, the researchers observed results that showed a Pearson correlation coefficient of 97% for kyphosis and 95% for lumbar lordosis [2]. This method involves some disadvantages, as it:

• requires special equipment including a motion analyzer and markers;
• requires the individual to be naked, which may not be the ideal option in all cultural contexts;
• requires background and proper lighting;
• lacks a specially designed software to be publically available;
• involves a high markers installation error, especially when they are set up by non-experts;
• has a non-inclusive algorithm, and it needs marker installation; an inclusive and comprehensive algorithm, nevertheless, could function without the use of the motion analyzer device and the markers.

However, with the increasing power of computers and the availability of mobile devices to people, new opportunities were created in the field of health and medicine. First, it helped to increase the knowledge of people and communities about health-related issues and keys to a healthy life. Second, new multidisciplinary branches in the field of health and medicine, namely bioinformatics, health informatics emerged. These fields relied on artificial intelligence and mathematical techniques as their basis for computation and reasoning. Until now, many approaches have been introduced to address the challenging task of diagnosing orthopedic diseases, and more specifically, spine disorders and vertebral defects. Methods in this area are roughly divided into two classes: Methods based on machine learning and conventional rule-based approaches.

### 4.1 Methods based on machine learning

The first group of approaches can grasp complex, non-linear relationships in the existing data [42]. [43] proposed a new approach based on random forests to diagnose Osteoarthritis and to provide an easy interpretation of results by using a continuous regression output. In work by [44], authors classified Osteoarthritis subjects using knee joints' sodium MRIs, which were transformed into radiant 3D acquisitions before and after the fluid suppression technique was applied. Every patient was then defined using 12 main features extracted from target image regions. [45] introduced a probabilistic boosting tree classifier in order to achieve a reliable segmentation approach to 3D vertebra CT spine images. Information on all vertebras shapes was exploited by using statistical shape modeling techniques that supported the segmentation process. In both [46] and [47], a weighted neighbor distance approach was manipulated along with a hierarchy combination of algorithms, which were first introduced by [48]. This approach represented morphology and was applied to the problem of Osteoarthritis diagnosis in MRIs of articular cartilage scans. In practice, this program was an image classifier that extracted global features from a training set of images, and discriminately assigned weights to them. In work by [49], an SVM based approach for determining needle entry site was developed. Authors applied midline detection, and template matching approaches to ultrasound spine images; in order to find the best classification features. In this research, more than 1,000 images were analyzed, and the accuracy degree of 95% on the training set and 92% on the test set were respectively obtained. A variation of the SVM model, named Least Squares SVM (LS-SVM), was manipulated by [50] in order to distinguish among scoliosis curve types. These curve types were obtained from 3D trunk images by using optical digitizers. When obtained, the 3D images were divided into horizontal slices. Then all the slices were broken down into patches, and extraction of geometric thoracic and lumbar descriptors was completed, which were later used as classification features after a thorough dimensionality reduction with Principal Component Analysis. In research by [51], a Computer-Aided Diagnosis system was developed to diagnose lumbar inter-vertebral discs degeneration anomaly automatically. The proposed method analyzed and extracted the features from both T1 and T2 weighted MR images in order to acquire different types of feature sets. A generative model was proposed by [52] to predict the progression and shape of idiopathic scoliosis affected spines, using 3D spine reconstruction of a set of X-Ray images. The researchers modeled multiple probabilistic structures, a geometric space with adequate structures and transform operators where high-dimensional data are reduced in size while retaining high-dimensional properties. A model based on deep learning and clustering techniques was proposed by [53] to detect Adolescent Idiopathic Scoliosis (AIS) automatically. In order to optimize the encoding-decoding of 3D spine model vectors, a neural network with an auto-encoder stack was trained. In two recent studies, [54] and [55] used the neural network, which was trained with ultrasound images for automatic detection of optimal vertebra level and Percutaneous spinal needle injection plane. [54] designed a neural network trained to partition ultrasound images into a multi-scale patch series recursively. In each iteration, Hadamard coefficients [56] were manipulated to transform standard wave-like ultrasound signatures into signatures of region-orientation correlations to identify distinguishing features of specific spinal patterns. Then input images were classified for both epidural and facet joint injections as either belonging to or not belonging to the target plane. In work reported by [55], a real-time scanner system was introduced. This scanner was implemented by using a convolutional neural network and a finite state transducer. The convolutional neural network was trained using a transfer learning methodology, where an inter-domain transfer of information was manipulated as a precondition for predicting accurately. Thus the convolutional neural network was able to detect and assign probabilities to three key positions along the patient's vertebral scanning – sacrum, intervertebral gaps, and vertebral bones. In a more recent neural-based approach by [57], two deep learning pipelines were employed; one for the cervical and the other for lumbar vertebra detection, using annotated clinical MRIs with information labels for each spinal vertebra as training data. Two CNNs were used for each pipeline for the identification of more general and specific vertebral features. Moreover, a component-based graphic model was built based on a layered graph to eliminate false positives and properly mark each vertebra. each layer of this graph shows the previously observed vertebrae and configurations of identified and marked vertebrae and measured the shortest path between them with the distance function based on mean and adjacent covariance matrices. Despite the ability of ML models, this capability is not the complete solution to any inquiry which a healthcare provider may make at the time of treatment. ML algorithms could overfit "predictions in false data correlations", as also noted by [58] and could identify predictors which "are not causes" despite their predictive power.

## 4.1 Rule-based approaches

For a long time, rule-based medical systems have been a reliable tool for health practitioners to validate their assumptions [59]. While providing a high degree of accuracy, these of type systems do not rely on any training data. The other advantage of a rule-based medical system, especially the medical rule-based expert system, is transparency, being easier to understand the reasoning from which conclusions are derived from. Until now, few rule-based works have been reported in the field of orthopedics. [60] believed the classification of cervical spine defects is the basis for surgical therapy and the recommendation for therapy. They reported an automatic rule-based classifier. This classifier required evidence from patient records, converted into a standardized form to facilitate data exchange and multimodal interoperability. [61] introduced a computer-based expert system to discover ergonomic risks for work-related musculoskeletal disorders. The proposed expert system utilized a knowledge base to classify risk factors into two central knowledge base units, in addition to four secondary knowledge base units that were related to the work environment and organization factors. This work suggested that a knowledge base system may be useful to assess a full-body assessment accurately. In a recent study, [62] developed an electronic health records(EHR) retrieval system. In particular, a retrieval system was introduced as a model for EHRs to support the medical decision-making process in managing spine defects using the information extracted from textual data on patient records. The patient cases were categorized according to classes of cervical spine defects, and the classification was based on the rules obtained from the relevant defect classification scheme. The classifier was applied to medical documents in a retrospective study. In [63], the authors presented a new 2-d segmentation, classification approach for structural changes in the hippocampal dendritic spines. An interactive rule-based module was leveraged for the classification of spines. Morphological features described essential attributes of the segmented spinal shapes; then, spines were categorized automatically into one of four classes: stubby, filopodia, mushroom, and spine head. A rule-based expert system for diagnosis of neck diseases was reported by [64]. The proposed expert system was based on forward chaining and decision trees inference; this system took user response in the form of YES/NO to complete the chain of inference. While lacking technical and scientific rigor, the authors claimed that their proposed approach is helpful in the diagnosis of neck pain diseases. Though many expert systems have been developed to help specialists with spine-related issues until now, no ES has been implemented for diagnosing and estimating the risk of development of spinal abnormalities.

## 5. Knowledge Engineering Method

The implementation of the proposed ES is displayed graphically in Figure 2. The cycles in Figure2 reflect the processes of level 1, and the indices next to them describe the processes of level 2, which are directly related to level 1 processes. In the proposed methodology, each process is repeated for any number of times until the medical expert system is formed in high quality and performance. The methodology is explained in this section.

questionnaire to detect swayback abnormality is represented in Table1 as a sample. Both of these are crucial parts of the knowledge acquisition process in the early stages.

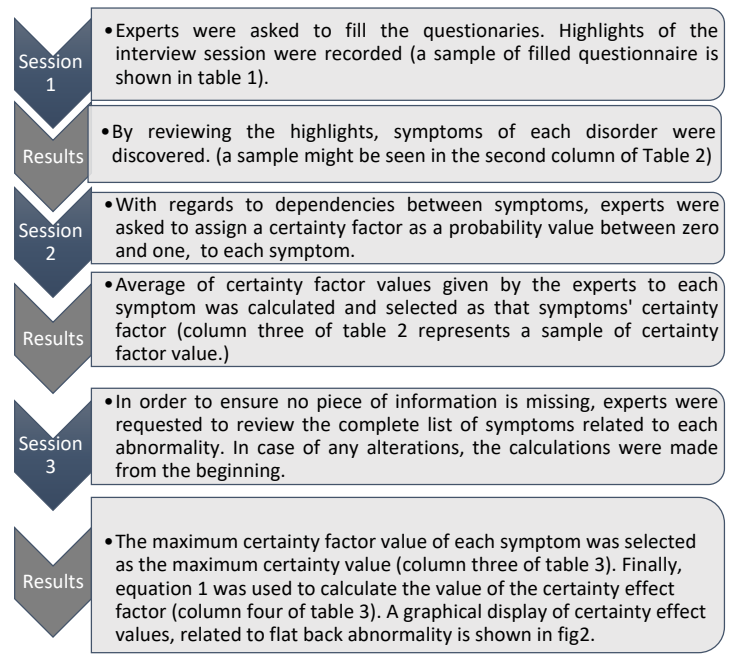

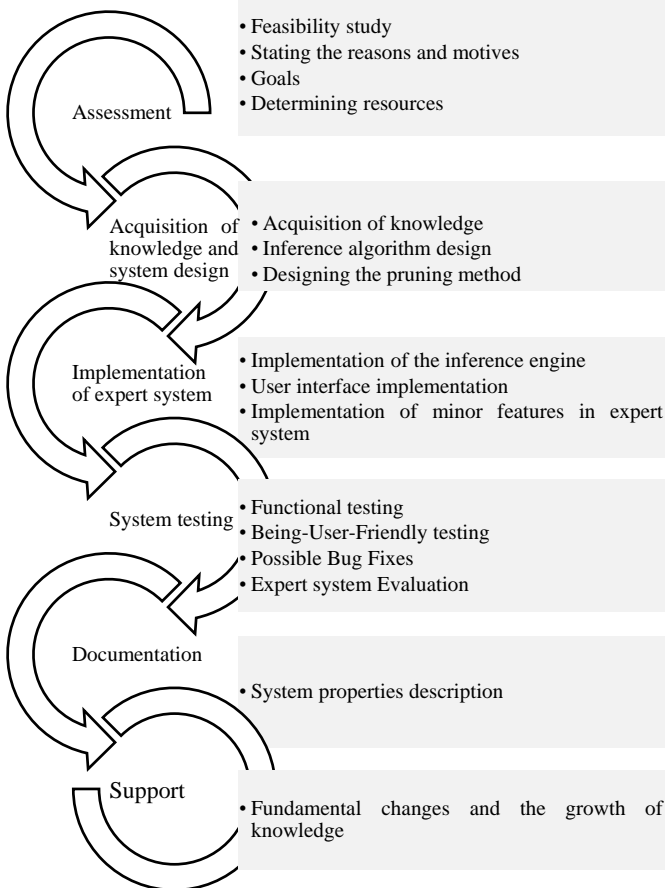

Fig. 2. Phases of knowledge engineering methodology to build the proposed expert system

Fig. 3. An overview of interview process

### 5.1 Knowledge Acquisition

The necessary resources of knowledge acquisition for system design included the following items:
1. four orthopaedists;
2. two physiotherapists;
3. a physical education instructor;
4. 63 articles, 18 reports, and four research projects, including all related signs and symptoms.

To design this system, the relevant knowledge for the detection and prevention of any deformities, as well as their symptoms and side effects, were identified by interviewing experts and using questionnaires. Figure 3 explains the interview process thoroughly; moreover, a filled

Table 1: sample of filled questionnaire by an expert in the first review session

| Type of Abnormality: |
|---|
| Round Back |
| Definition: |
| Round Back is a spinal abnormality in which an excessive outward curve of the spine results in an abnormal rounding of the upper back. Typically, the thoracic spine should have a natural kyphosis between 20 to 45 degrees, while postural or structural abnormalities can result in a curve that is beyond this range. Patients of all ages might be diagnosed with Round Back. Nevertheless, this condition is common during adolescence, when the bones rapidly grow. |
| Diagnosis: |
| 1-Wall Test 2-Shortness of Hamstring Muscle test 3-checkerboard 4-plumb-line 5-Spinal mouse 6-x-rays 7-MRI scan |
| Side Effects: |
| 1- Intervertebral disk herniation 2- Pain in Trapezius muscle and Chest 3-Disk space narrowing 4-Pelvic obliquity 5-Muscle imbalance 6-Fatigue 10-Tear or wear of the lumbar (lower) spine 11-Loss of height 12-Mild to severe back pain 13-Difficulty standing straight upright 14-Limited physical functions 15-Digestive problems 16-Body image problems 17-Breathing problems |
| Causes: |
| 1-Degeneration of adjacent intervertebral discs 2-Overstretched or weak upper back muscles 3-Fractures 4-Osteoporosis 5-Scheuermann's disease 6-Imheritence and birth defects 7-Inheritance 8-Syndromes 9- Cancer and cancer treatments 10-Tuberculosis 11-Paget's disease 12- Spina bifida 13-Muscular dystrophy 14-Neurofibromatosis |

Next, the experts (orthopaedists) were asked to classify the symptoms based on their importance and rate of occurrence in patients with the spinal anomaly. After the classification of symptoms, experts assigned a unique certainty factor value to each symptom. This value determines the degree of certainty of observing a symptom related to an anomaly in unhealthy patients (as seen in Table 2). Finally, the certainty factors assigned by the experts were summed, and the average value was considered to be a single certainty factor. In the third column of Table 3, the final certainty factor can be seen. This factor was derived from the stages mentioned above. Similarly, after classifying the symptoms associated with each other, the certainty effect factor in the fourth column of Table 3 was obtained through maximizing the certainty factor of the symptoms via Formula 3. This factor denotes the $CF$ of the rule premises and serves as a threshold to confirm the observance of symptom in a patient Figure 4 schematically illustrates certainty effect values related to flat back anomaly.

$$certainty\ effect = \frac{maximum\ certainty\ factor\ in\ class}{total\ certainty\ factors}$$
(3)

Table 2. Sample of symptoms and related factors of flat back abnormality

| # | Symptoms | Certainty from the perspective of an expert | Probability | Cumulative probability | probability amendment | Symptoms class |
|---|---|---|---|---|---|---|
| 1 | Flat thoracic spine (loss of natural curve in the upper back) | 80% | 0.355 | 0.355 | 0.644 | A |
| 2 | Flat lumbar spine (loss of natural low back curve) | 60% | 0.266 | 0.622 | 0.733 | A |
| 3 | Reduced flexibility of spine | 30% | 0.133 | 0.755 | 0.866 | B |
| 4 | Loss of intervertebral disk space | 20% | 0.088 | 0.848 | 0.911 | C |
| 5 | Degenerative disk disease | 20% | 0.088 | 0.933 | 0.911 | D |
| 6 | Compression fractures | 10% | 0.044 | 0.977 | 0.955 | E |
| 7 | Digestive problems | 5% | 0.022 | 1 | 0.977 | F |

Table 3. Sample of the certainty effect factor calculated using the certainty factor in Table 2 for the flat back anomaly

| # | Symptoms class | Maximum certainty from the perspective of an expert in class | Certainty effect | Uncertainty effect | Cumulative certainty effect |
|---|---|---|---|---|---|
| 1 | A | 80% | 0.484 | 0.515 | 0.484 |
| 2 | B | 30% | 0.181 | 0.818 | 0.666 |
| 3 | C | 20% | 0.121 | 0.878 | 0.787 |
| 4 | D | 20% | 0.121 | 0.878 | 0.909 |
| 5 | E | 10% | 0.060 | 0.939 | 0.969 |
| 6 | F | 5% | 0.030 | 0.969 | 1 |

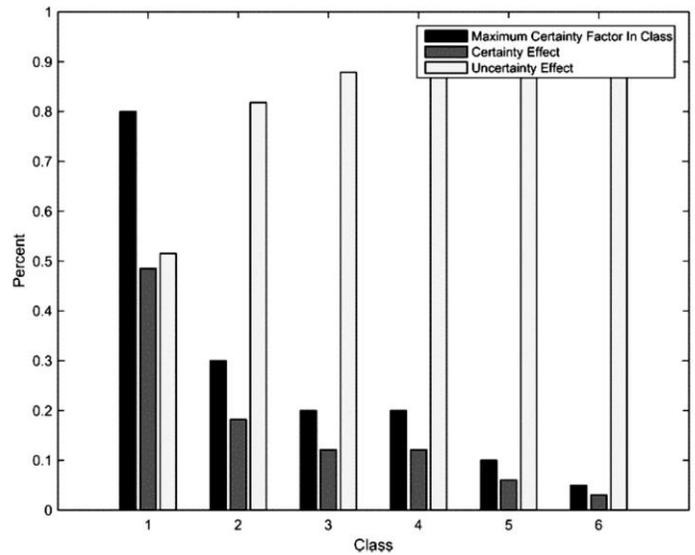

Fig. 4. A sample of certainty effect factor for classified symptoms of the flat back anomaly

### 5.2 The Proposed Algorithm of Inference Engine

After knowledge acquisition, inference methods and decision-making processes were determined based on existing knowledge. To emulate experts' thinking and to avoid the redundant occurrence of the rules, a hybrid algorithm was formulated for inference to obtain an excellent performance in terms of completeness, optimality, time complexity, and space complexity. The algorithm was an incorporation of the BCI [32] and Uncertainty [37]. The most frequently used method of the uncertainty principle in the MESs relies on the certainty factor [38]. However, in the proposed hybrid algorithm, another factor, known as the "certainty effect factor" was used. Applying this factor in the BCI would make it possible for the system to function without storing users' trail questions and answers. Moreover, in every stage of the decision-making process, one could provide the feedback of statistical results by knowing the previous step of questions and answers. Besides, one could *prune* the chain through a binary decision tree. Because ESs, especially medical diagnosis systems, are, in most cases, multi-agent, dynamic, inaccessible, uncertain, and continuous, the method proposed in this study drew on the certainty effect factor. This factor helped to simplify the system design by integrating symptoms into some classes and by removing redundant symptoms across the classes. This algorithm frames a "certainty memory" which functions very similarly to *the human thinking* process. The certainty memory sorts the rules in descending order according to the value of their symptoms' certainty effect factor at each moment. The inference process begins when the user specifies the type of abnormality he/she wants to be diagnosed. Next, a question is asked, and the user responds to that question by entering a numerical value within a range of 0 and 100. If the input value is equal or higher than the present certainty effect value of the symptom in certainty memory, it is concluded that examined symptom is observed in the patient with a high degree of certainty. Then the system will remove the original value of that symptom's "certainty effect" value. Next, the user input value of that symptom will be replaced as the new "certainty effect factor" while the premise is satisfied. However, If the user response is any value smaller than the existing value, the predefined "certainty effect" value remains unchanged. That is due to the low chances of observing that symptom in the patient, and as a result, that premise will not be satisfied. Afterward, all the satisfied premises are matched with their antecedents, and goal-related rules are fired. As soon as the rules are fired, the certainty factor (CF) of each rule is calculated based on its' premises. At each stage, when a new rule is fired, the average sum of certainty factor of rules will be calculated as "certainty degree" of the chain of inference. Equation 4. is used to calculate this factor, which represents the likelihood of existence or risk of developing an anomaly in a person. While delivering a good level accuracy, this approach made efficient reasoning possible. The proposed inference algorithm and approximate responses would require fewer rules than the conventional method of the decision tree and explicit responses. The flowchart of this algorithm can be seen in Figure 5.

$$Certainty\ Degree = \frac{\sum_{i=1}^{n} CF(i)}{n}$$
(4)

The advantages of this inference algorithm are as follows:
- In the proposed algorithm, the user's questioning-answering process can be stopped at any point, while reaching an acceptable answer at any stage of the decision-making process, because the events are independent.
- There is no need to store the path because the certainty memory can compute the relevant statistics at any stage.
- The possibility to return to previous questions in this algorithm can be accomplished.

### 5.3 Implementation of knowledge base and user interface

After knowledge acquisition was completed, and the inference algorithm was developed, knowledge engineers implemented the knowledge base, using Java Drools Library. On the other side, system designers must realize that ordinary users are not software engineers or knowledge engineers who can work in a complex environment of an ES. There is no doubt that systems with an inconvenient interface cannot be functionally efficient in actual practice. In this study, after the knowledge base was deployed, the graphical interface was designed by JavaFX. It has communicational features such as text-to-speech tool. Figure 6 illustrates the proposed ES during the questioning-answering process initiated by the user. During the evaluation, all the users could conveniently work with this system. The customized version of the ES is compatible with these operating systems: Windows XP, Windows 7, Windows 8, Windows 10, Windows Server, Windows Vista, Mac, Linux, and Open Solaris.

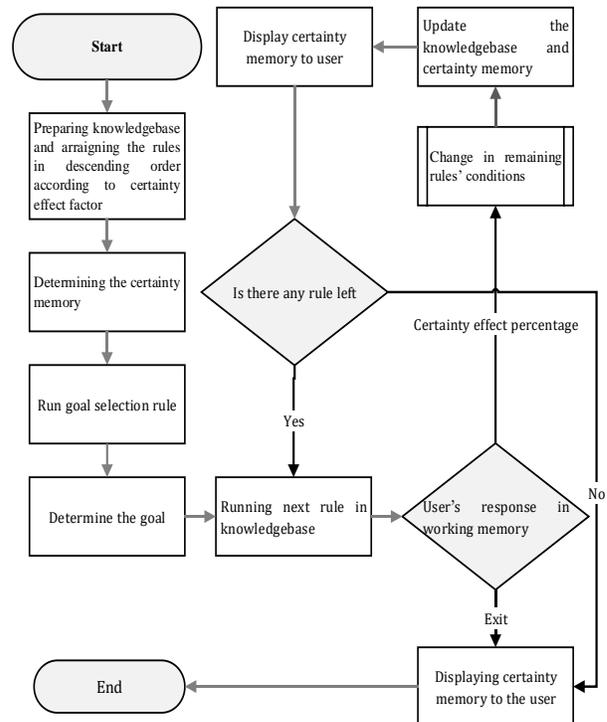

Fig. 5. Flowchart of inference engine algorithm

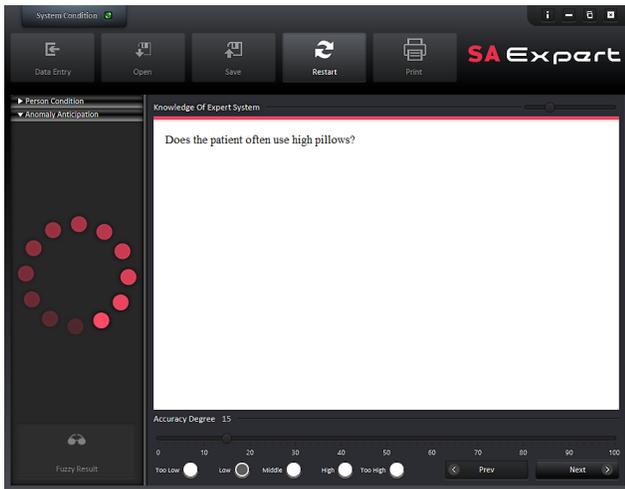

Fig. 6. A sample view of the proposed expert system during question-answering process

### 5.4. Diagnosis in practice

The developed expert system relies on the degree of certainty in a series of responses made by the user to determine the order of questions that have to be asked. After each response, user input is taken into account as a certainty amount, and the system state is updated. Then the system provides the most suitable question to the user. The question answering process ends at the point where no related question is left in the working memory. A flow of questions and answers to diagnose scoliosis abnormality in a 31-year-old female patient is shown in table 4. Note that the patient was diagnosed with scoliosis abnormality after a close check-up by experts.
.

Table 4: sample of question answering to diagnose scoliosis abnormality

| # | Questions | User Input/Degree of Certainty |
|---|---|---|
| 1 | What is the patient's sex? | female |
| 2 | How old is the patient? | 17 |
| 3 | How tall is the patient? (in cm) | 160 |
| 4 | How much does the patient weight? | 65 |
| 5 | To calculate weekly physical activity, answer following questions: Average heart rate during activity: Average daily physical activity Average weekly physical activity | 115 low low |
| 6 | Does the patient's spine has a sideways curve? | 89% |
| 7 | Does the patient have an "S"- or "C"-shaped spine? | 97% |
| 8 | Does one shoulders look higher than the other? | 89% |
| 9 | Does one shoulder blade stick out more than the other? | 90% |
| 10 | Does the patient usually carry heavy objects using only one hand? Or play tennis or throw spears? | 0% |
| 11 | Are the hips uneven? | 66% |
| 12 | Does the plumb line not pass within 1.7 cm of the posterosuperior corner of the S1 vertebral body ? | 88% |
| 13 | Is one of the legs longer than the other? | 97% |
| 14 | Are the abdominal muscles weak? | 94% |
| | Result: Patient is diagnosed with scoliosis abnormality with certainty degree of 89% | |

### 6. Results

The Developed expert system was evaluated in two separate phases, In the first phase, participants or real-world samples (patients diagnosed with spinal abnormalities and also healthy people) were used to examine the accuracy of the system. Whereas in the second phase, four orthopaedists evaluated the system using medical records of patients suffering from spinal disorders; based on their knowledge and experience. Both phases are discussed as follows.

### 6.1 Phase 1: Evaluation based on real-world samples

Two random statistical samples (groups of participants) were selected to evaluate the accuracy of the proposed ES results. Four orthopaedists were asked to introduce the ES to the participants, to gain feedback about the functioning of the ES. The first sample included individuals with spinal anomalies, while the second class of samples comprised of people who had a healthy spine. The evaluation process at this stage included predicting the likelihood of existence or chance of development of a spinal disorder in an individual. In doing so, 400 individuals were selected to form the samples. One hundred twenty of them were already diagnosed with one of the spinal anomalies by the experts, while 280 of them seemed to be in reasonable condition. Each expert examined only 30 samples from the group diagnosed with spinal disorders and 70 samples from a healthy group of individuals. Every one of the experts ran the system as a total number of 30 times for unhealthy samples, and five times for each healthy sample in the second group. The reason behind repeating the question-answering process for the healthy samples was to determine the likelihood of the development of any of the five spinal disorders in that patient in the future. This could help the experts to prevent the development of spinal disorders in healthy patients by providing precautionary actions or giving the necessary tips to them. Finally, the system was run for the total number of 1520 executions, and results were gathered. Table 5 shows the statistical information about the participants with anomalies, and Table 6, represents the information related to healthy/normal individuals. $\bar{X}$, $S$ and $CV$ denote mean or average probability, standard deviation and coefficient of variation respectively. As Table 5 represents, evaluation results show a minimum average accuracy value of 0.923 for the correct detection of cervical lordosis in unhealthy patients. Also, the system was successful in true detection of kyphosis disorder with a maximum average accuracy value of 0.940 in samples group suffering from the spinal anomaly. Table 6 demonstrates the mean value of the likelihood of developing any of the five anomalies in healthy samples. Figures 7 and 8 show the error bar for the detection of anomalies in both healthy and unhealthy groups of participants.

Table 5. The results of expert system in examining the sample with anomalies

| # | Anomaly name | $\bar{X}$ | $S$ | $CV$ |
|---|---|---|---|---|
| 1 | Scoliosis | 0.935 | 0.060 | 0.065 |
| 2 | Flat back | 0.952 | 0.085 | 0.089 |
| 3 | Kyphosis. | 0.946 | 0.037 | 0.039 |
| 4 | Cervical lordosis | 0.923 | 0.034 | 0.037 |
| 5 | Swayback | 0.930 | 0.022 | 0.024 |

Table 6. The results of expert system in examining the sample without anomalies

| # | Anomaly name | $\bar{X}$ | $S$ | $CV$ |
|---|---|---|---|---|
| 1 | Scoliosis | 0.071 | 0.045 | 0.634 |
| 2 | Flat back | 0.027 | 0.022 | 0.814 |
| 3 | Kyphosis | 0.042 | 0.022 | 0.523 |
| 4 | Cervical lordosis | 0.057 | 0.036 | 0.631 |
| 5 | Swayback | 0.069 | 0.045 | 0.652 |

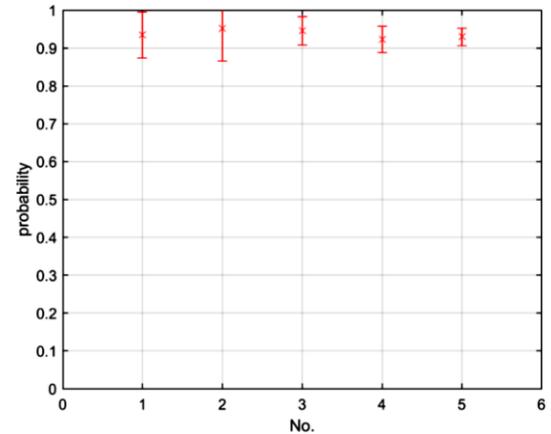

Fig. 7. The error bar, and phase1 evaluation results of the expert system for the samples with anomalies

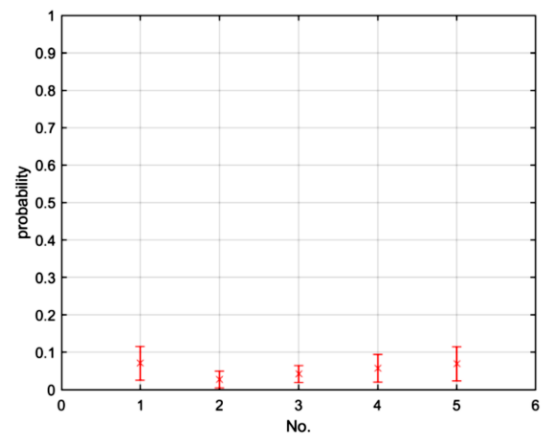

Fig. 8. The error bar, and phase1 evaluation results of expert system for the samples without anomaly

### 6.2 Phase 2: Evaluations based on medical records

Four orthopaedists participated in evaluating the system. Experts were provided with medical records of 2000 patients diagnosed with spinal abnormalities in earlier examinations. The records were comprised of 400 instances per each anomaly. All the records were segmented and placed under the related abnormality class. For each spinal disorder, 100 cases were selected by each expert. Experts were then asked to review the given medical records. Features of these cases were later fed to ES as inputs through question and answering process. Each orthopaedist ran the system 100 times for each anomaly, as a sum of 400 executions by four experts for each anomaly and total executions of 2000 for all the anomalies. Experts determined to what extent the results of the expert system are accurate in the real world applications, and how properly it can detect an abnormality.

As Table 7 and Figure 9 show, Evaluation results from the four experts are very close, with minimum average accuracy value of 0.915 for true detection of cervical lordosis and swayback, and maximum average accuracy of 0.940 for true detection of kyphosis spinal disorder.

**6.3 Cut-off values in evaluation procedure**

When the four experts completed the two phases of evaluation, they were asked to give their feedback about the system performance. Based on their opinions, there are three different situations: first when a sample is correctly diagnosed, second when a sample needs closer examination, third when a given sample is healthy. Then experts were asked to describe their degree of certainty about system's accuracy in each of these three cases in terms of numerical values. Table 8 shows a complete list of cut-off values for detection of spinal abnormalities which were obtained based on the experts' feedback. Threshold values where categorized into three classes:

1-TPD or True Positive Detection: the minimum cut-off value showing that an examined sample is definitely suffering from any type of spinal disorders.
2-TND or True Negative Detection: the maximum cut-off value showing that a given sample is healthy.
3-RFI or Requirement of Further Examinations: any value lower than TPDs or higher than TNDs is regarded as a RFI, suggesting that a sample with diagnosis probability within this span of numbers requires close examinations.

For each anomaly the estimated average of values given by experts was considered as final threshold value.

**6.4 Discussion**

Experts participating in the evaluation of the system unanimously agreed over the integrity of the knowledge and validity of assessments. However, the proposed work is still subject to a few limitations. First, while experts found the evaluations to be reliable, they still argued over the degree of correctness of the results. For example, once the expert system detected scoliosis abnormality with a minimum certainty of 76% in a sample, all the four orthopaedists were confident that the examined sample was suffering from scoliosis. Meanwhile, there are situations where the system cannot detect abnormalities with a high degree of accuracy. For instance, in the examination of patients, where scoliosis abnormality had been diagnosed with accuracy degrees between 50% and 75%, at least one out of four experts believed further physical examinations are required before reaching a final judgment. This condition is very similar to the real-world cases where an orthopaedist fails to make a definitive diagnosis and starts further investigations. At this stage, the health specialist may seek or ask for additional details and information. In cases where abnormalities were detected with certainty degree equal or below 50%, all experts agreed that the patient was healthy; nonetheless, that certainty value still reflected the risk of developing anomaly in the examined sample.

Knowing that experts are prone to errors, primarily due to prolonged activity, the purpose of this research was to create an expert system in the form of intelligent software to help experts diagnose spinal anomalies more accurately. Evaluations show that the proposed approach was successful in detecting spinal anomalies and to assess the risk of anomaly development.

Table 7: Evaluation results of the system by experts

| # | Anomaly Name | Expert No.1 | Expert No.2 | Expert No.3 | Expert No.4 | $\overline{X}'$ | S | CV |
|---|---|---|---|---|---|---|---|---|
| 1 | Scoliosis | 0.853 | 0.948 | 0.957 | 0.961 | 0.925 | 0.050 | 0.054 |
| 2 | Flat back | 0.860 | 0.975 | 0.933 | 0.944 | 0.925 | 0.046 | 0.050 |
| 3 | Kyphosis | 0.923 | 0.911 | 0.963 | 0.975 | 0.940 | 0.029 | 0.031 |
| 4 | Cervical lordosis | 0.880 | 0.939 | 0.942 | 0.910 | 0.915 | 0.026 | 0.028 |
| 5 | Swayback | 0.891 | 0.916 | 0.911 | 0.954 | 0.915 | 0.025 | 0.027 |

Table 7: Cut-off values estimated based on experts' feed back

| # | Anomaly Name | Expert No.1 | Expert No.2 | Expert No.3 | Expert No.4 | $\overline{Cut-off}$ |
|---|---|---|---|---|---|---|
| | True Positive Detection | | | | | |
| 1 | Scoliosis | 0.725 | 0.740 | 0.770 | 0.805 | 0.760 |
| 2 | Flat back | 0.655 | 0.810 | 0.770 | 0.785 | 0.755 |
| 3 | Kyphosis | 0.790 | 0.780 | 0.815 | 0.835 | 0.785 |
| 4 | Cervical lordosis | 0.675 | 0.750 | 0.810 | 0.725 | 0.740 |
| 5 | Swayback | 0.680 | 0.745 | 0.735 | 0.820 | 0.745 |
| | True Negative Detection | | | | | |
| 1 | Scoliosis | 0.445 | 0.500 | 0.515 | 0.540 | 0.500 |
| 2 | Flat back | 0.445 | 0.530 | 0.455 | 0.510 | 0.485 |
| 3 | Kyphosis | 0.485 | 0.470 | 0.550 | 0.575 | 0.520 |
| 4 | Cervical lordosis | 0.395 | 0.495 | 0.510 | 0.460 | 0.465 |
| 5 | Swayback | 0.420 | 0.475 | 0.455 | 0.530 | 0.470 |

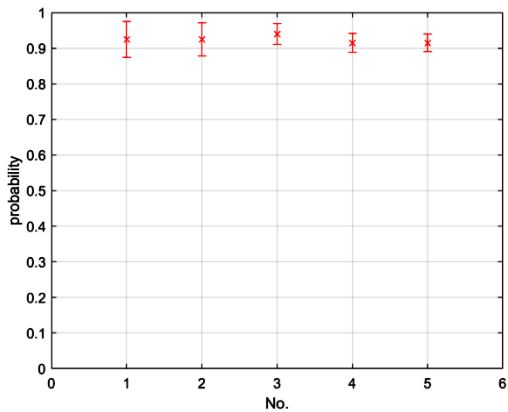

Fig. 9. The error bar, and phase2 evaluation results of the system for each anomaly by experts

## 7. Conclusion and Future Studies

This study proposed a novel, non-invasive method for detecting and preventing spinal anomalies. The expert system designed and implemented in this research can be used as a reliable assistant by medical experts, specially orthopaedists. This expert system helps specialists to validate the accuracy and integrity of results of spinal disorders diagnosis. It may also serve as an intelligent educational software for medical students to learn the process of spinal disorder diagnosis, and discover the symptoms and variables related to each spinal anomaly. By suggesting an optimal algorithm for the inference engine to emulate experts' inference, the proposed method could successfully reason based on an optimal number of rules in a flexible question-answering process. It could also bring about advantages in comparison with the malfunctioning of sole use of conditional probabilities. In the conducted evaluation, the system was able to successfully detect anomalies in an individual by using anthropometric dimensions, everyday life habits, anatomy, and physiological data. The extensive use of this expert system by specialists may help a more accurate diagnosis and preventing the spread of spinal anomalies, leading to a healthier life for the people of the world. The present research may be further improved by using fuzzy inference or neural networks. Besides, other factors of anthropometric dimensions of healthy participants, anomalies, and medical information by researchers may be incorporated to yield better results.

## Abbreviations

AIS: Adolescent Idiopathic Scoliosis; BCI: Backward Chaining Inference; CF: Certainty Factor; CNN: Convolutional Neural Network; CT: Computed tomography; EHR: Electronic Health Records; ES: Expert System; LS: Least Squares; MES: Medical Expert System; ML: Machine Learning; MR: Magnetic Resonance; MRI: Magnetic resonance imaging; RFI: Requirement of Further Examinations; SVM: Support Vector Machine; TPD: True Positive Detection; TND: True Negative Detection;

## Ethics approval and consent to participate
Not Applicable.

## Consent for publication
Not Applicable.

## Availability of data and material
The data that support the findings of this study are available from *Tehran University of Medical Sciences* but restrictions apply to the availability of these data, which were used under license for the current study, and so are not publicly available. Data are however available from the authors upon reasonable request and with permission of *Tehran University of Medical Sciences*.

## Competing Interest
The authors declare that they have no competing interests.

## Funding
Not Applicable.

## Author's contribution
*Seyed Mohammad Sadegh Dashti* and *Seyedeh Fatemeh Dashti* conceived of the presented idea. *Seyed Mohammad Sadegh Dashti* developed the theoretical formalism. *Seyed Mohammad Sadegh Dashti* and *Seyedeh Fatemeh Dashti* accomplished the process of knowledge acquisition. Both *Seyed Mohammad Sadegh Dashti* and *Seyedeh Fatemeh Dashti* developed the inference algorithm and encoded the knowledge to the knowledge-base. *Seyed Mohammad Sadegh Dashti* verified the analytical methods. *Seyed Mohammad Sadegh Dashti* and *Seyedeh Fatemeh Dashti* designed and developed the user interface. *Seyed Mohammad Sadegh Dashti* and *Seyedeh Fatemeh Dashti* supervised carrying the experiments out. *Seyedeh Fatemeh Dashti* performed the analytic calculations and performed the numerical simulations. *Seyed Mohammad Sadegh Dashti* and *Seyedeh Fatemeh Dashti* discussed the results. *Seyedeh Fatemeh Dashti* wrote the whole manuscript with support from *Seyed Mohammad Sadegh Dashti*.

## Acknowledgments
Authors would like to greatly appreciate Dr. Azita Yazdani from the Department of Health Information Management, Tehran University of Medical Sciences, for providing the medical records without which this research could not be evaluated.

## 8. References

[1] Jibril, I., Agajo, J.; **Development of a Medical Expert System for Hypertensive Patients Diagnosis: A Knowledge-Based Rules**; Advances in Electrical and Telecommunication Engineering (AETE); 2018.

[2] Chaudhuri S.B., Rahman M; **Design of a Medical Expert System (MES) Based on Rough Set Theory for Detection of Cardiovascular Diseases**. Progress in Advanced Computing and Intelligent Engineering. Advances in Intelligent Systems and Computing; 2018.

[3] AsadKarami S., Ghasemi G.; **The effect of eight weeks of NASM exercises on Sway back of high school female students**; The Scientific Journal of Rehabilitation Medicine; 2018.

[4] HeidariMoghaddam R; **Reviewing the role of major thalassemia major in spinal abnormality development**; Research journal of medical college; 2013.

[5] Tolouei A; **Developing an expert system to detect blood cancer**; Journal of Health Management; 2010.

[6] Jibril I. Z., Agajo J., Ajao L. A., Kolo J. G. & Inalegwu, O. C; **Development of a Medical Expert System for Hypertensive Patients Diagnosis: A Knowledge-Based Rules**; Advances in Electrical and Telecommunication Engineering 1; 2018.

[7] Elaine N. Marieb.; **Human Anatomy & Physiology**; Pearson Education, Inc; 2018;

[8] Khoubi M; **Corrective and therapeutic exercises for treatment of spinal abnormalities**; Physical Sciences publication; September 2015.

[9] Keshvari F, Mirdarikvandi M; **Applicable Corrective Cxercises** ; Sokhanvaran publicatons; 2018.

[10] Daneshmandi M, Sedaghati P; **New corrective approaches to treatment of Parkinson's disease**; Boshra publications; December 2017.

[11] Machida M.; **Neurological Research in Idiopathic Scoliosis;** Pathogenesis of Idiopathic Scoliosis; 2018.

[12] Mao, S., Qian, B; **Quantitative evaluation of the relationship between COMP promoter methylation and the susceptibility and curve progression of adolescent idiopathic scoliosis**; European Spine Journal; 2018.

[13] Akazawa, T., Kotani, T; **Midlife changes of health-related quality of life in adolescent idiopathic scoliosis patients who underwent spinal fusion during adolescence**; European Journal of Orthopaedic Surgery & Traumatology; 2018.

[14] Nanjundappa S., Harshavardhana MD; **Results of Bracing for Juvenile Idiopathic Scoliosis**; Spine Deformity; 2018.

[15] Jason Brumitt; **Core Assessment and Training;** Human Kinetics; 2010.

[16] Ilbeigi S, Mehrshad N, Afzalpour ME, Yousefi M; **Diagnosing spinal disorders by using markers mounted on spinous excrescences, sports medicine**; 2010.

[17] Daneshmandi H., Pourhosseini H., SardarMA.; **Analyzing spinal disorders among boys and girls**; Journal of motion; 2004.

[18] Uei, H., Tokuhashi, Y; **Multiple vertebral fractures associated with glucocorticoid-induced osteoporosis treated with teriparatide followed by kyphosis correction fusion: a case report**; Osteoporosis International; 2018.

[19] Kobets, A.J., Komlos, D.; **Congenital cervical kyphosis in an infant with Ehlers-Danlos syndrome**; Child's Nervous System; 2018.

[20] Schule r Thomas C; **Segmental Lumbar Lordosis: Manual Versus Computer-Assisted Measurement Using Seven Different Techniques** ; J Spinal Disord Tech ; 2004.

[21] Cressey, Eric; **Strategies for Correcting Bad Posture – Part 4**; EricCressey.com; 2014.

[22] Robertson PA, Armstrong WA; **Lordosis Re-Creation in TLIF and PLIF: A Cadaveric Study of the Influence of Surgical Bone Resection and Cage Angle**; Spine (Phila Pa 1976); 2018.

[23] Sayari A, Farahani A, Ghanbarzadeh M; **Review and Comparison of Structural and Aerobic Corrective Exercises Affecting Pulmonary Function of Students with Kyphosis**; Olympics Journal; 2007.

[24] Ghafouri F; **Relation of Kyphosis with Depression and Anxiety Among Athletic and Non-Athlete Male Students in Tehran Universities**; Journal of Research in Sport Science; 2007.

[25] MirBagheri R, Ghasemi B; **Effect of Eight Weeks of Massage Therapy on Lumbar Lordosis**; Second Conference on applicable researches in Sport Science; 2018.

[26] Mohammadi Sh, Mokhtarinia HR; **Investigating the Effects of Different Working Postures on Cognitive Performance**. Archives of Rehabilitation; 2018.

[27] Choubineh A, Moudi A; **Human-Design and Ergonomics**; PhisNet, Tehran; 2015.

[28] Smith S, Kandel A; **Verification and validation of rule-based expert systems**; CRC Press; 2018.

[29] Ignizio, James P; **Introduction to expert systems: the development and implementation of rule-based expert systems**; 1991.

[30] Giarratano, Joseph C; **Gary Riley; Expert systems;** PWS publishing co.; 1998.

[31] Mohammadi Motlagh H.A., Minaei Bidgoli B.; **Design and implementation of a web-based fuzzy expert system for diagnosing depressive disorder**; Applied Intelligence; 2018.

[32] Al-Ajlan A; **The comparison between forward and backward chaining**; International Journal of Machine Learning and Computing.; 2015.

[33] Grosan, Crina, and Ajith Abraham.; **Rule-based expert systems**; In Intelligent Systems; 2011.

[34] Caignya A., Coussementa K.; **A new hybrid classification algorithm for customer churn prediction based on logistic regression and decision trees**; European Journal of Operational Research; 2018.

[35] A. Jose., A. Prasad, **Design and Development of a Rule Based Expert System for AACR: A Study of the Application of Artificial Intelligence Techniques in Library and Information Field**; VDM Verlag, Saarbrücken, Germany, 2011.

[36] M. Sagheb-Tehrani; **Expert systems development: Some issues of design process**; ACM SIGSOFT Software Engineering Notes, vol. 30,no. 2, pp. 1-5, 2005.

[37] [Shapiro EY; **Logic Programs With Uncertainties: A Tool for Implementing Rule-Based Systems;** In IJCAI; 1983.

[38] Baudryab G., Macharis C.; **Range-based Multi-Actor Multi-Criteria Analysis: A combined method of Multi-Actor Multi-Criteria Analysis and Monte Carlo simulation to support participatory decision making under uncertainty**; European Journal of Operational Research; 2018.

[39] Fon GT, Pitt MJ, Thies AC; **Thoracic Kyphosis: range in normal subjects**; Am J Roentgenol; 134: 979-983; 1980.

[40] Mahmoudi F, Shahrjerdi S; **Changes in Pain and Kyphosis Angle Following a Corrective Exercise Program in Elderly Women: A Randomized Controlled Trial**. JRUMS; 2017.

[41] Kado DM, Prenovost K, Crandall C; **Narrative review: Hyperkyphosis in older persons**; Ann. Intern. Med. 147 (5); 330-8; 2007.

[42] Chen JH, Asch SM.; **Machine learning and prediction in medicine—beyond the peak of inflated expectations**; The New England journal of medicine; 2017.

[43] Kotti M, Duffell LD, Faisal AA, McGregor AH; **Detecting knee osteoarthritis and its discriminating parameters using random forests**; Medical engineering & physics; 2017.

[44] Madelin G, Poidevin F, Makrymallis A, Regatte RR.; **Classification of sodium MRI data of cartilage using machine learning. Magnetic resonance in medicine;** 2015.

[45] Mirzaalian H, Wels M, Heimann T, Kelm BM, Suehling M.; **Fast and robust 3D vertebra segmentation using statistical shape models**; In2013 35th annual international conference of the IEEE engineering in medicine and biology society (EMBC); 2013.

[46] Ashinsky BG, Coletta CE, Bouhrara M, Lukas VA, Boyle JM, Reiter DA, Neu CP, Goldberg IG, Spencer RG.; **Machine learning classification of OARSI-scored human articular cartilage using magnetic resonance imaging**; Osteoarthritis and cartilage; 2015.

[47] Ashinsky BG, Bouhrara M, Coletta CE, Lehallier B, Urish KL, Lin PC, Goldberg IG, Spencer RG.; **Predicting early symptomatic osteoarthritis in the human knee using machine learning classification of magnetic resonance images from the osteoarthritis initiative**; Journal of Orthopaedic Research; 2017.

[48] Shamir L, Orlov N, Eckley DM, Macura T, Johnston J, Goldberg IG.; **Wndchrm–an open source utility for biological image analysis**; Source code for biology and medicine; 2008.

[49] Yu S, Tan KK, Sng BL, Li S, Sia AT.; **Lumbar ultrasound image feature extraction and classification with support vector machine**; Ultrasound in medicine & biology; 2015.

[50] Adankon MM, Dansereau J, Labelle H, Cheriet F.; **Noninvasive classification system of scoliosis curve types using least-squares support vector machines**; Artificial intelligence in medicine; 2012.

[51] Oktay AB, Albayrak NB, Akgul YS.; **Computer aided diagnosis of degenerative intervertebral disc diseases from lumbar MR images**; Computerized Medical Imaging and Graphics; 2014.

[52] Kadoury S, Mandel W, Roy-Beaudry M, Nault ML, Parent S.; **3-D morphology prediction of progressive spinal deformities from probabilistic modeling of discriminant manifolds**; IEEE transactions on medical imaging; 2017.

[53] Thong W, Parent S, Wu J, Aubin CE, Labelle H, Kadoury S.; **Three-dimensional morphology study of surgical adolescent idiopathic scoliosis patient from encoded geometric models**; European Spine Journal; 2016.

[54] Pesteie M, Abolmaesumi P, Ashab HA, Lessoway VA, Massey S, Gunka V, Rohling RN.; **Real-time ultrasound image classification for spine anesthesia using local directional Hadamard features**; International journal of computer assisted radiology and surgery; 2015.

[55] Hetherington J, Lessoway V, Gunka V, Abolmaesumi P, Rohling R.; **SLIDE: automatic spine level identification system using a deep convolutional neural network**; International journal of computer assisted radiology and surgery; 2017.

[56] Beauchamp KG.; **Applications of Walsh and related functions: with an introduction to sequency theory**; Academic press; 1984.

[57] Forsberg D, Sjöblom E, Sunshine JL.; **Detection and labeling of vertebrae in MR images using deep learning with clinical annotations as training data**; Journal of digital imaging; 2017.

[58] Obermeyer Z, Emanuel EJ.; **Predicting the future—big data, machine learning, and clinical medicine**; The New England journal of medicine; 2016.

[59] Abu-Nasser B.; **Medical Expert Systems Survey**; International Journal of Engineering and Information Systems (IJEAIS); 2017.

[60] Deng Y, Groll MJ, Denecke K.; **Rule-based Cervical Spine Defect Classification Using Medical Narratives**; Studies in health technology and informatics; 2015.

[61] Pavlovic-Veselinovic S, Hedge A, Veselinovic M; **An ergonomic expert system for risk assessment of work-related musculo-skeletal disorders**; International Journal of Industrial Ergonomics; 2016.

[62] Deng Y, Denecke K.; **Patient Records Retrieval System for Integrated Care in Treatment of Cervical Spine Defect**; InVLDB Workshop on Data Management and Analytics for Medicine and Healthcare; 2016.

[63] Basu S, Plewczynski D, Saha S, Roszkowska M, Magnowska M, Baczynska E, Wlodarczyk J.; **2dSpAn: semiautomated 2-d segmentation, classification and analysis of hippocampal dendritic spine plasticity**; Bioinformatics; 2016.

[64] Naser SS, ALmursheidi SH.; **A Knowledge Based System for Neck Pain Diagnosis**; World Wide Journal of Multidisciplinary Research and Development (WWJMRD); 2016.